\documentclass[preprint,12pt]{elsarticle}

\usepackage{amsmath}
\usepackage{siunitx}
\usepackage{multirow}

\journal{Medical Image Analysis}

\begin{document}

\begin{frontmatter}



\title{SurgeMOD: Translating image-space tissue motions into vision-based surgical forces}


\author[kings,cair]{Mikel De Iturrate Reyzabal \corref{cor1}}
\author[kings,cair]{Dionysios Malas}
\author[cair]{Shuai Wang}
\author[kings]{Sebastien Ourselin}
\author[cair,kings]{Hongbin Liu}

\affiliation[kings]{organization={Department of Biomedical Engineering and Imaging Science, King's College London},
            addressline={1 Lambeth Palace Rd}, 
            city={London},
            postcode={SE1 7EU}, 
            country={United Kingdom}}

\affiliation[cair]{organization={Centre for Artificial Intelligence and Robotics, Chinese Academy of Science},
            addressline={3/F,17W, Science Park West Avenue, Hong Kong Science Park, Pak Shek Kok}, 
            city={Hong Kong},
            state={New Territories},
            country={SAR Hong Kong}}

\cortext[cor1]{Corresponding author: {mikel.de\_iturrate\_reyzabal@kcl.ac.uk}}

\begin{abstract}
We present a new approach for vision-based force estimation in Minimally Invasive Robotic Surgery based on frequency domain basis of motion of organs derived directly from video. Using internal movements generated by natural processes like breathing or the cardiac cycle, we infer the image-space basis of the motion on the frequency domain. As we are working with this representation, we discretize the problem to a limited amount of low-frequencies to build an image-space mechanical model of the environment. We use this pre-built model to define our force estimation problem as a dynamic constraint problem. We demonstrate that this method can estimate point contact forces reliably for silicone phantom and ex-vivo experiments, matching real readings from a force sensor. In addition, we perform qualitative experiments in which we synthesize coherent force textures from surgical videos over a certain region of interest selected by the user. Our method demonstrates good results for both quantitative and qualitative analysis, providing a good starting point for a purely vision-based method for surgical force estimation.

\end{abstract}


\end{frontmatter}


\section{Introduction}
\label{sec:intro}

Minimally Invasive Robotic Surgery (MIRS) has become the standard approach for specific surgical procedures in urology \cite{mikhail_urologic_2020} and gynecology \cite{moon_robotic_2020}, due to their safety and reduced recovery times \cite{wee_systematic_2020}. Moreover, their use in more complex cardiac surgical procedures \cite{kypson_robotic_2004} such as endoscopic coronary artery bypass \cite{tabaie_endoscopic_1999} has been demonstrated in the last years. However, current robotic systems for MIRS present some limitations, with the lack of real-time haptic feedback being one of the most meaningful examples.

To compensate for the lack of tactile feedback, surgeons use the observable deformations of the tissue in contact with the tool to inherit the applied forces. However, they usually require some previous knowledge of the workspace to determine elastic properties and the governing physical constraints of the system. As the surgical environment is in constant motion from the natural processes inside the body such as breathing \cite{frazier_impact_2004}, heartbeat \cite{bello_deep-learning_2019} and peristalsis \cite{burns_peristaltic_1967}, surgeons take advantage of these cyclic movements to obtain this information before the actual surgery. In summary, a combination of both gives a complete understanding of the scene in a two step learning fashion.

\paragraph{\textbf{Videos as textures}} Certain moving scenarios can be conceptualized as dynamic textures \cite{doretto_dynamic_2003}, wherein videos are modeled as space-time samples of a stochastic process. These textures can represent smooth, natural motions such as flames, moving trees or waves; and they have been commonly used for video segmentation, classification or encoding \cite{casas_4d_2014, chan_mixtures_2005, chan_classifying_2007, doretto_dynamic_2003}. Using such techniques also allows videos to be synthesized from single reference frames, mimicking the motion of known environments \cite{chuang_animating_2005}. In recent years, with the appearance of generative diffusion models these techniques have also been used to generate these textures based on a data-driven approach \cite{li_generative_2023}. Our method makes use of this approximation of dynamic textures to build a model for the basis of motion of soft tissue prior to surgery and, therefore, learn the underlying basis of the 2D motion of the environemnt.

\paragraph{\textbf{Soft tissue motion}} The motion of soft tissue under an external force is a highly studied area of research, as under the solid mechanics approximation, both variables are mathematically related with a 2nd ODE \cite{bathe_finite_2006, omar_review_2022}. The main approach in this area is to use 3D representations of the motions. This methodology uses organ meshes approximated from pre-operative image modalities such as CT or MRI. In order to solve the small deformation approach formula, we load these models on a physic solvers such as SOFA \cite{allard_sofa_2007} or MuJoCo \cite{todorov_mujoco_2012} that use numerical methods to approximate the solution of the system. However, such approaches rely on knowledge of the mechanical properties of the tissue for an accurate estimation of the motion. Nevertheless, work has been done on a more data-driven approach, in which information from a video was used to approximate these parameters \cite{mountney_three-dimensional_2010, giannarou_probabilistic_2013, stoyanov_soft-tissue_2005}. Although they present promising results, they still require  additional processing to calculate the point cloud based on stereo-vision disparity maps.

In addition, the use of modal analysis \cite{shabana_theory_1991, shabana_constrained_1991} as a representation of motion in the medical domain has been investigated recently \cite{hauser_interactive_2003, basdogan_haptics_2004}. This methodology has been largely validated in other areas of engineering \cite{huang_interactive_2011, james_dyrt_2002} to analyse the viability of structures and determine the 3D basis of motion of systems. In contrast to previous work, our method learns the tissue basis of motion directly from video and represents them in the image-space inspired by the work from Davis et al. \cite{davis_image-space_2015}.

\paragraph{\textbf{Force estimation in Minimally Invasive Robotic Surgery}} 

Reliable estimation of forces is a key factor in the success of a surgery. Previous work on deep learning vision-based force estimation field required additional multi-modal inputs such as robot kinematics or additional force readings to obtain well-founded predictions \cite{chua_toward_2021}. Additionally, they require specific time sampling, kinematic chains dimensionality reduction or more complex visual encoders to marginally improve the results of previous research \cite{ko_vision-based_2023, lee_toward_2021, haouchine_vision-based_2018, jung_vision-based_2021, mikel_dafoes_2024}. Consequently, most of the methods in this area lack  theoretical background that gives a deeper overview on the underlying  working principle. However, there has been some work on the area regarding the use of optical flow for the estimation of forces \cite{neidhardt_optical_2023}, but still requires  volumetric data inputs. Therefore, our method presents a pure computer vision problem based on the theoretical approach derived from Modal Analysis. In addition, we provide forces in the observable image-space rather than in the world coordinates, and they can be delivered as a single mean vector of the contact force or mapped to a desired region of interest.

\section{Theory}
\label{sec:overview}

Consider a region $\boldsymbol{\psi}$ of a soft elastic tissue in an image $I_{0}$ that expands over $N$ degrees of freedom. We can define this region as $\boldsymbol{\psi} = (\mathbf{p}_{1}, \mathbf{p}_{2}, \text{...}, \mathbf{p}_{N})$. Under small deformations, we characterize the movement of this area over a certain period of time $T$ as a set of texture maps known as \textit{motion textures}, introduced at \cite{chuang_animating_2005}. These maps are defined as $\mathcal{M}_{\boldsymbol{\psi}} = \left\{M_{t}(\boldsymbol{\psi})| \ t = 1, 2, \text{...}, T\right\}$ maps the original values of the region $I_{0}(\boldsymbol{\psi})$ to the ones at a given time instance $t$:

\begin{equation}
\label{eq:motion-texture}
    I'_{t}(\boldsymbol{\psi} + M_{t}(\boldsymbol{\psi})) = I_0(\boldsymbol{\psi})
\end{equation} 

where $I'_{t}$ is the forward-wrapped image at a time instance $t$.

\begin{figure}[t]
    \centering
    \includegraphics[width=\textwidth]{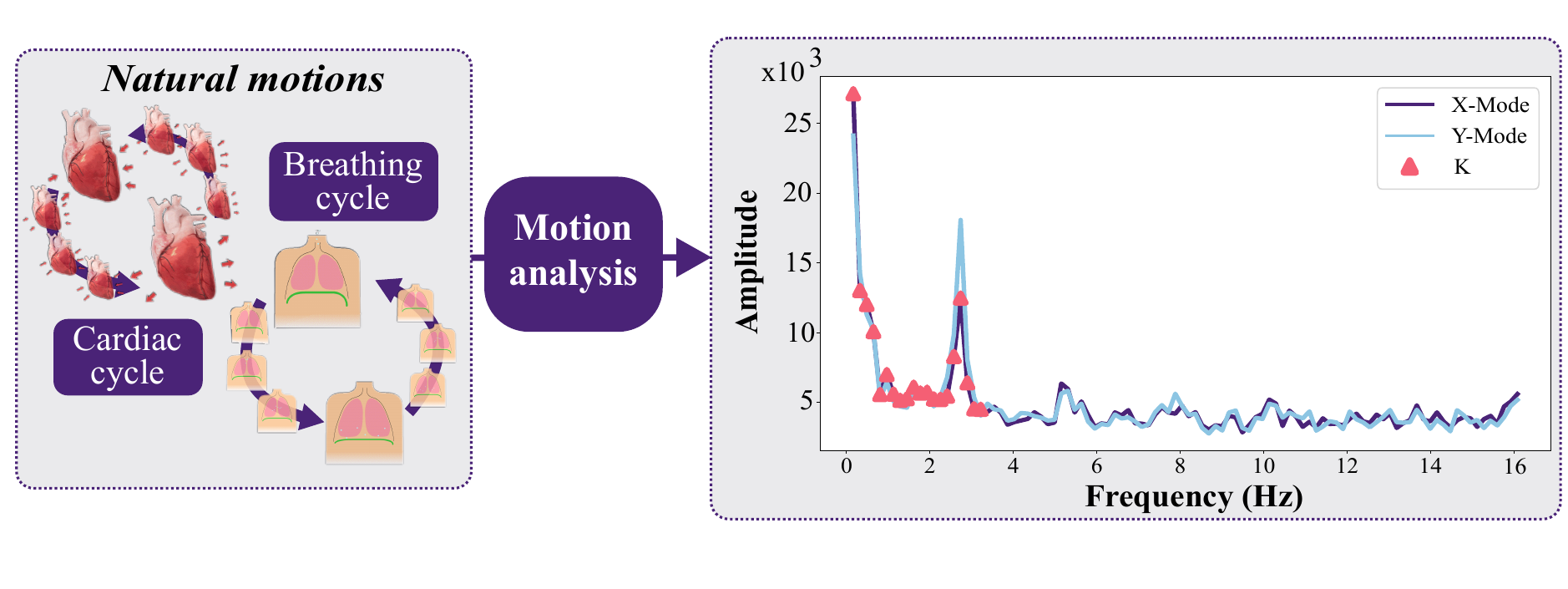}
    \caption{The graph presents the mean amplitude of the frequency analysis of tissue motion on camera for every frequency over the time period $T$. We analyse all the positive frequencies of the spectrum $T/2$. The selected $K$ frequencies are marked using the pink triangular markers. From the graph, we can observe that the majority of the information from the motion generated by the natural cycles on the left is accumulated on the first $15-20$ frequencies.}
    \label{fig:power_spectrum}
\end{figure}

As we are working with soft tissue, we can model these \textit{motion textures} as an accumulation of mass-spring systems with varying frequencies and phases \cite{mollemans_tetrahedral_2003, omar_review_2022, duan_volume_2016}. Due to their sinusoidal behavior, it is common to analyse such systems in the frequency domain, using Modal Analysis \cite{shabana_theory_1991, bathe_finite_2006}. Hence, as demonstrated by Davis et al. \cite{davis_image-space_2015}, under the assumption of a well-spaced frequency domain, these spectral maps represent the image space basis of motion of the captured object. Therefore, we can build the so called \textit{mode shapes} \cite{li_generative_2023} as the set of spectral maps $\mathcal{S}_{\boldsymbol{\psi}} = \left\{S_{\omega}(\boldsymbol{\psi}) |\ \omega = 0, 1, \text{...}, {T}/{2} - 1\right\}$ at a given set of frequencies $\omega$. By definition, the \textit{mode shapes} and the \textit{motion texture} are directly related by the Fast Fourier Transform:

\begin{equation}
\label{eq: fft}
    \mathcal{S}_{\boldsymbol{\psi}} = \mathbf{FFT}(\mathcal{M}_{\boldsymbol{\psi}})
\end{equation}

One of the most interesting properties of Modal Analysis is that it provides a direct relationship between frequencies and applied forces. This helps to define problem specific constraints, such as setting a cutoff limit for the frequency, as high-frequency modes are only achieved under big excitation which are not present during surgical procedures. For example, natural motions of winds or trees usually happen at low frequencies \cite{ota_1fsup_2003, diener_wind_2009}. Therefore, we could conclude that forces during a surgery are usually around $0-5$\si{N}\cite{bahar_surgeon-centered_2020, basdogan_haptics_2004}, meaning that we can set an upper limit for the frequency. In consequence, we can limit our analysis to a set of frequencies $K$, that follows the assumption $T >> K$.

In order to experimentally define $K$, we analyse the mean power spectrum of the mode shapes across all the collected videos presented at Sec. \ref{sub:data}. We can see from Fig. \ref{fig:power_spectrum} that the values for the spectrum exponentially drop after a few initial frequencies. Considering both, the literature research and the analytical approach, we decide to set $K=20$, this means that we are taking the frequencies from $0.2-3\si{Hz}$. We discard the first frequency as it represents the rigid body motion, translation and rotation, which does not directly relate to the deformation of the soft tissue.

\section{Materials and Methods}

\begin{figure}[t]
    \centering
    \includegraphics[width=\textwidth]{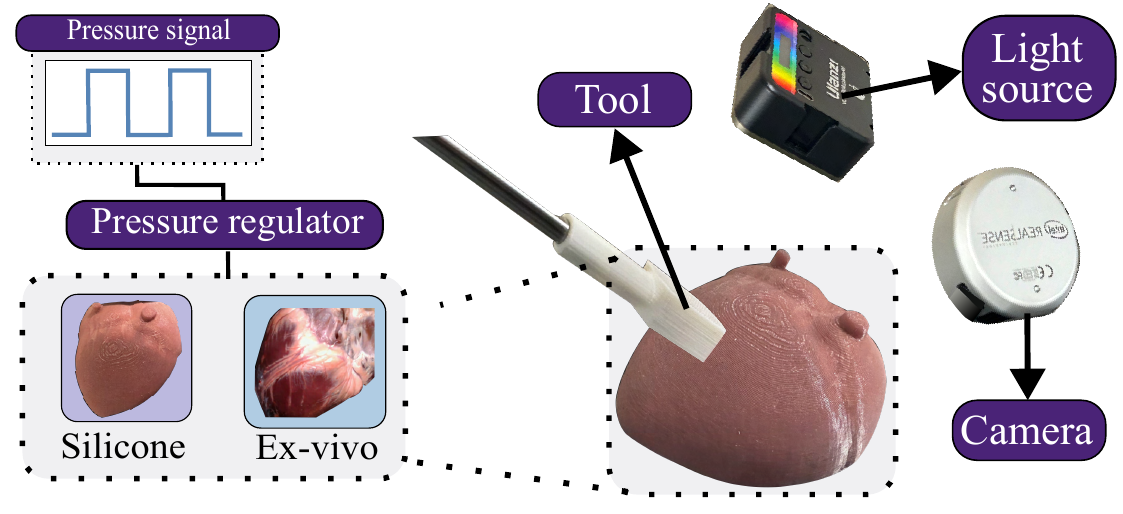}
    \caption{Graphical diagram of our experimental setup. Our testing platform is either: silicone phantom or an ex-vivo porcine cardiac tissue. This is connected to a pressure regulator that controls the cyclic cardiac motion of the heart. We place the camera close enough to obtain a clear recording of the motion and a light source to create consistent lighting to reduce its effect on the optical flow prediction.}
    \label{fig:exp-setup}
\end{figure}

\subsection{Surgical environment modelling \& data collection}
\label{sub:data}

Access to real-time force readings from a real surgical video is limited. Therefore, to test the viability of our force estimation methodology presented in this research, we prepare a silicone phantom and ex-vivo evaluation platform. Our phantom, made of  DragonSkin20 silicone, mimics the heart shape and elastic characteristics, and our ex-vivo platform is formed by a procine heart. Additionally, for our silicone phantom, we add some superficial texture using a mix of different colors to add traceable features for our algorithm. We emulate the heart pumping motion at a frequency of $2$\si{Hz} (close to $60$\si{bpm}, as the cardiac cycle is divided into two phases) using a proportional pressure regulator to generate a pressure-square wave (0.15\si{bar}).

Using a finger like tool, we poke the environments and perform a set of experiments which are explained in detail in Section \ref{sub:experiments}. During each of our experiments, we record both the video and the forces of the interactions. For the video, we use the RGB sensor of an Intel Realsense L515 camera, placed $15$\si{cm} away from the tissue. To record the force, we use a  M3815A1 6-axis F/T sensor from Sunrise Instruments, position between the poking tool and the robot arm end effector. The tool for the interaction is a PLC piece of 10mm diameter, mimicking the geometry of a "finger". To link the tool and the Elfin 3 Hans Robot arm we use a $5$\si{mm} diameter metallic rod of $150$\si{mm} length. We cover all the experimental setup using a box and we use a direct light source to avoid unwanted artifacts generated by inconsistent lightning.

\subsection{Experiments}
\label{sub:experiments}

We collect a total of 30 videos, 15 for each of the environments (silicone phantom and ex-vivo real porcine tissue described in \ref{sub:data}). In addition, we use real surgical videos of natural body motions from the Hamlyn Centre Laparoscopic / Endoscopic Video Datasets \cite{mountney_three-dimensional_2010, giannarou_probabilistic_2013, stoyanov_soft-tissue_2005}. These videos include motion of a liver and a diaphragm due to breathing, and cardiac motion.

We divide the collected data into two different sources of information: \textbf{natural body motions} under no external influence and \textbf{tool-tissue interactions}, tissue motion due to external forces. The first set consists of baseline videos with no external forces, that give our method the necessary information to compute the mode shapes mentioned in Sec. \ref{sec:overview}. On the other hand, the second set of experiments is formed by 4 different scenarios. For these scenarios, we collect both visual and force sensor data. The main objective of this set is to determine the force using solely visual cues. Therefore, in order to analyse different type of interactions, we divide the experimental objectives into:

\begin{itemize}
    \item \textbf{Force scale:} Keeping a consistent camera position and contact point, touch the phantom with different force intensities.
    \item \textbf{Shear force:} Analyse the effectiveness of the method detecting the force over a surface when the normal force onto the surface is smaller than the shear force. Move around the phantom close to the surface, generating a deformation field.
    \item \textbf{Force positioning}: Keeping the amount of force constant, touch at different point on the phantom. Detect the different reactions from the phantom to the touch.
    \item \textbf{Breathing noise:} Keeping the dynamic behavior of the environment, touch the tissue on the same position with different intensities. Detect the force scale and the effect of breathing motion onto the force prediction.
\end{itemize}

\begin{figure}[t]
    \centering
    \includegraphics[width=0.9\textwidth]{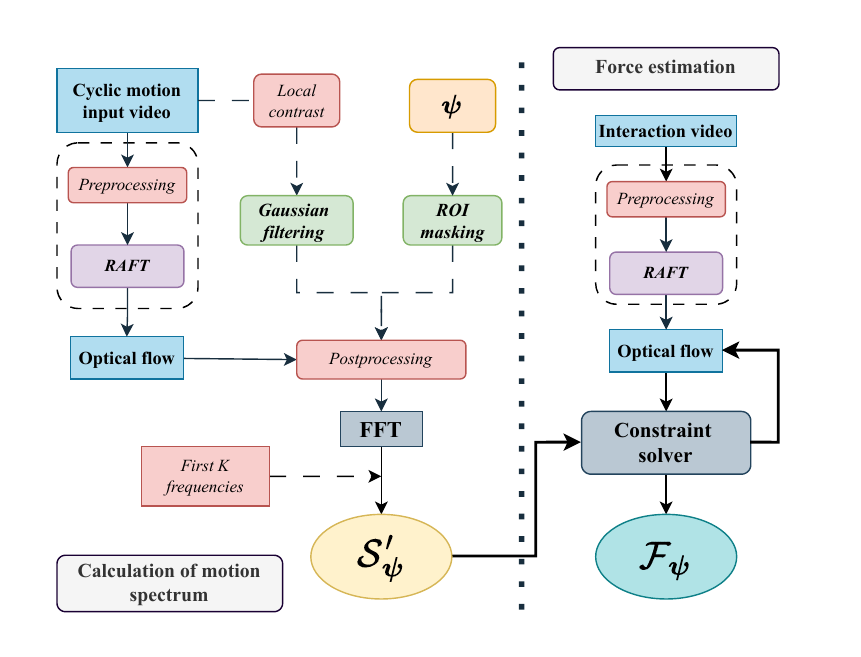}
    \caption{Diagram representation of our proposed force estimation algorithm. The force estimator runs in two different steps: calculation of the mode shape (left) and force estimation (right). The optical flow postprocessing helps to have a more uniform distributions of the power in the frequency domain, as it can be observed at Fig. \ref{fig:power_spectrum}}
    \label{fig:algorithm}
\end{figure}

\subsection{Algorithm: Image-space surgical force calculation}

In this section we present our algorithm for the vision-based force calculation from a surgical video. The process is divided into two parts: the mode shape calculation and the estimation of forces.

\subsubsection{Mode shape calculation and selection}
The calculation of the complex mode shapes $S'_{\omega}$ is done in different steps. In order to generalize the solution of this task we decided to use foundation models as the main prediction tool. We first determine the optical flow from the input video, as shown in the diagram from Fig. \ref{fig:algorithm},. For this, we use the RAFT module \cite{teed_raft_2020}, as it is the most generalized approach for such tasks. Consequently, to fit our input batch to model requirements, we add a pre-processing (resize and normalization) pipeline.

In order to occlude the motions generated by unclear or occluded regions, we filter the output optical flow. We use local contrast or Gaussian filtering based on the current RGB frame, enhancing the motions from highly visible parts and masking out the motion from barely visible areas. As we want to focus on the motion from the organ, we use a binary mask based on the reference frame for the object. This isolates the pixels corresponding to the organ, similar to the process of creating a 3D mesh of the desired surgical environment. Finally, we calculate the Fast Fourier Transform of the processed optical flow. In order to simplify the problem and, as we previously discussed in Sec. \ref{sec:overview}, we isolate the first $20$ frequencies from the original mode shapes $S_{\omega}$, resulting in the modified $S'_{\omega}$ that we use for the rest of our research.

\subsubsection{Force estimation: The dynamic constraint problem}

For this section, we are trying to determine the force in a surgical video during the time period $T'$ on our region of interest $\boldsymbol{\psi}$. Let as defined the force texture as the collection of 2D maps that gives its value related with the motion texture, with mathematical notation $\mathcal{F}_{\boldsymbol{\psi}} = \left\{F_{t}(\boldsymbol{\psi}) | t = 1, \text{...}, T' \right\}$. We assume that this force generates a visible change on the structure and surface of the tissue and, also, that it is the main contributor to the movement between two consecutive time instances. Focusing at a given time $t$, by the definition of the \textit{motion textures} the desired displacement and force are $M^{*}_{t}(\boldsymbol{\psi})$ and $F_{t}(\boldsymbol{\psi})$, respectively. Under previously mentioned assumptions, we can denote the constraint equation as \cite{hauser_interactive_2003}:

\begin{equation}
\label{eq:acc-cons}
    \ddot{M}^{*}_{t}(\boldsymbol{\psi}) = \mathcal{S}'_{\boldsymbol{\psi}} \left(\ddot{M}_{t-1}(\boldsymbol{\psi}) + \mathcal{S}'^{T}_{\boldsymbol{\psi}} F_{t}(\boldsymbol{\psi})\right)
\end{equation}

where $\ddot{M}_{t-1}$ is the acceleration of the motion texture in the previous time step. Solving Equation \ref{eq:acc-cons} for the force.

\begin{equation}
\label{eq:f-acc-cons}
    F_{t}(\boldsymbol{\psi}) = \left(\mathcal{S}'_{\boldsymbol{\psi}} \mathcal{S}'^{T}_{\boldsymbol{\psi}}\right)^{-\mathbf{P}} \left( \ddot{M}^{*}_{t}(\boldsymbol{\psi}) - \mathcal{S}'_{\boldsymbol{\psi}} \ddot{M}_{t-1}(\boldsymbol{\psi})\right)
\end{equation}

where $-\mathbf{P}$ is the Moore-Penrose pseudoinverse. The velocity constraint transforms the force from Equation \ref{eq:f-acc-cons} in a force impulse $\Delta t F_{t}(\boldsymbol{\psi})$, where $\Delta t$ is the rate of frame update from the camera ($\Delta t = 1/\si{fps}$), giving

\begin{equation}
\label{eq:f-v-cons}
    F_{t}(\boldsymbol{\psi}) = \frac{1}{\Delta t} \left(\mathcal{S}'_{\boldsymbol{\psi}} \mathcal{S}'^{T}_{\boldsymbol{\psi}}\right)^{-\mathbf{P}} \left( \dot{M}^{*}_{t}(\boldsymbol{\psi}) - \mathcal{S}'_{\boldsymbol{\psi}} \dot{M}_{t-1}(\boldsymbol{\psi})\right)
\end{equation}

For the displacement constraint, we consider the force to be an impulse again, but we need to add a motion correction factor $\mathcal{C}$ to correct for the amount of motion generated by each of the independent frequencies for the given sampling time $\Delta t$. We modify Equation \ref{eq:f-v-cons} to

\begin{equation}
    F_{t}(\boldsymbol{\psi}) = \frac{1}{\Delta t ^ {2}} \left(\mathcal{S}'_{\boldsymbol{\psi}} \mathcal{C} \mathcal{S}'^{T}_{\boldsymbol{\psi}}\right)^{-\mathbf{P}} \left( M^{*}_{t}(\boldsymbol{\psi}) - \mathcal{S}'_{\boldsymbol{\psi}} M_{t-1}(\boldsymbol{\psi})\right)
\end{equation}

The algorithm for the force calculation follows a similar strategy as the calculation of the mode shapes. We present a diagram on how both of the algorithm are combined at Fig. \ref{fig:algorithm}.

\section{Results}

Based on the expeirments presented in Sec.\ref{sub:experiments}, we perform different analysis to cover the different applications of our methodology. These contain a set of quantitative and qualitative results. The first involves the calculation of the contact forces as the weighted sum of forces over the area. The latter collects the synthesis of the force textures $\mathcal{F}_{\boldsymbol{\psi}}$ and force signal stabilization using the motion prior correction from the mode shapes.

\begin{figure}
    \centering
    \includegraphics[width=0.95\textwidth]{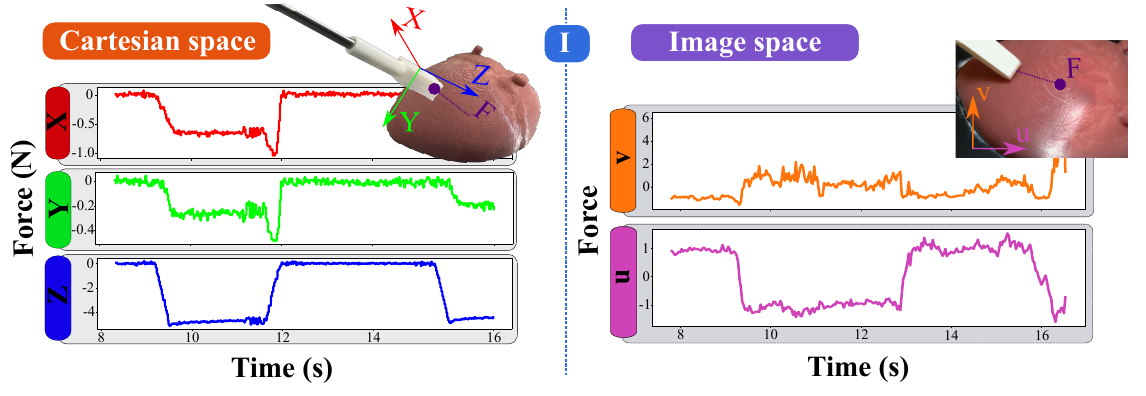}
    \caption{Representation of the different spaces in which forces are represented. On the left we have the 3D Cartesian space representation of the force. This is the force recorded by the sensor for our testing data. On the right, we have the projection of the 3D space into the camera coordinates u-v. The force is also projected into this image space and it is related to the Cartesian space through the intrinsic parameters of the camera.}
    \label{fig:force-projection}
\end{figure}

\subsection{Quantitative analysis}

\begin{figure}[t]
    \centering
    \includegraphics[width=0.95\textwidth]{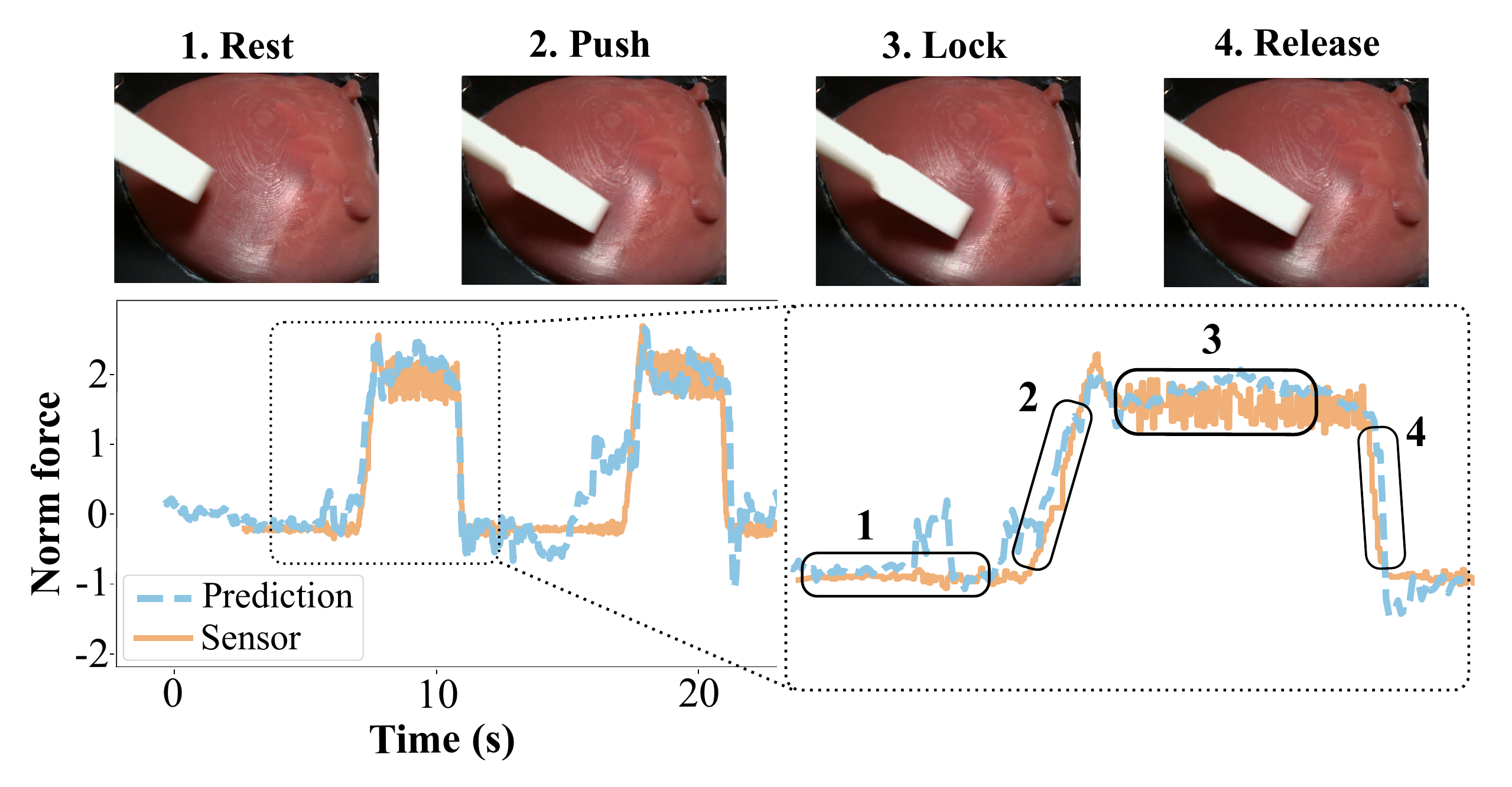}
    \caption{Results of the force comparison between the predicted and measured contact force. This example presents a classic poking scenario and we can divide it into four different steps based on the force response: 1) rest, there is no contact; 2) push, the contact starts and increases in intensity; 3) lock, the moment on maximum force in which we keep the tool for a short period of time; and 4) release, the contact between the tool and the tissue is relaxed.}
    \label{fig:force-comparison}
\end{figure}

In this section, we compare our methodology against the contact forces measured by the force sensor, $F_{s}$. For this study, we use our original force prediction methodology to predict the mean force close to the contact. We select a small region of interest $\boldsymbol{\psi}$ in the reference frame where the contact is taking place. As this region has multiple pixels, in order to match the force sensor readings with our estimated values, we take the weighted sum of the predicted force texture over the region of interest. We define this contact at a given time instance $t$ as $\hat{F}_{t}(\boldsymbol{\psi})$ in Equation \ref{eq:contact-force},

\begin{equation}
\label{eq:contact-force}
    \hat{F}_{t}(\boldsymbol{\psi}) = \frac{1}{w_{\boldsymbol{\psi}} h_{\boldsymbol{\psi}}} \sum_{i = 1}^{N} F_{t}(\mathbf{p}_{i})
\end{equation}

where $w_{\boldsymbol{\psi}}$ and $h_{\boldsymbol{\psi}}$ are, respectively, the width and the height of the region of interest.

As we have both forces represented in different reference spaces, we need to project the recorded forces into the image-space. An example of how we interpret both forces can be seen in Figure \ref{fig:force-projection}. We use the camera orientation to first rotate the tool force to the world coordinates from the camera. We then use the camera intrinsic parameters to project the force in the image-space, so they are comparable 
 with our predicted forces. For a better comparison, we normalize both the predicted and recorded force by subtracting the mean and dividing by the standard deviation. Results with an example are shown in Figure \ref{fig:force-comparison}.

As the force sensor and the video are recorded at different sampling rates, we use the maximum value of the normalized cross-correlation between the real and the predicted signal. We use this metric to validate our method across all the different videos and scenarios. First, we resample the real label to be the same size as the prediction. Then, we calculate the normalized cross correlation value using the definition in Equation \ref{eq:cross-corr},

\begin{equation}
\label{eq:cross-corr}
    R_{\hat{F}F^{s}}^{\text{max}} = \text{max}_{k} \left( \frac{\sum_{t=1}^{T'} \hat{F}_{t} \cdot F_{s}(t+k)}{\sqrt{\sum_{t=1}^{T'} \hat{F}_{t}^2 \cdot \sum_{t=1}^{T'} F_{s}(t+k)^2}} \right)
\end{equation}

where $k$ is the lag of the signal and accounts for the human problems during the recordings of the experimental data. We use the optimal lag of this maximum normalized cross-correlation to calculate two additional metrics: cosine similarity and mean absolute error. The results for both the phantom and ex-vivo experiments can be found in Table \ref{tab:metrics} at the end of this paper.

For a more complete vision on our algorithm we compared both directions of the force in the image space $\mathbf{u}(\rightarrow)$ and $\textbf{v}(\uparrow)$ independently; in addition to the norm value of the force. 

Moreover, we analyze each of the environment silicone and ex-vivo independently for each of the force types discussed on the experimental design. These forces are normal and shear forces; the results can be found at Tables \ref{tab:phantom} and \ref{tab:ex-vivo}.

\begin{table}[t]
\label{tab:metrics}
\caption{Validation metrics for our custom experimental setup. CC refers to cross-correlation, RMSE refers to root mean square error, CS refers to cosine similarity and MAE refers to mean absolute error.}
\hspace{1.5em}
\small
\centering
\begin{tabular}{|l|c|l|ccc|}
\hline
\multicolumn{3}{|c|}{\textbf{Environments}} & \textbf{Complete} & \textbf{Phantom} & \textbf{Ex-Vivo} \\ 
\hline
\multirow{6}{*}{\textbf{CC}} & \multirow{2}{*}{\textbf{Norm}}   & \textit{Mean \& Std}      & 0.832 $\pm$ 0.068 & 0.82$\pm$0.059    & 0.866$\pm$0.079    \\
& & \textit{Max - Min} & 0.935 - 0.671 & 0.903 - 0.671      & 0.935 - 0.754      \\
\cline{2-6}
& \multirow{2}{*}{\textbf{u} $(\rightarrow)$} & \textit{Mean \& Std} & 0.63 $\pm$ 0.188 & 0.647 $\pm$ 0.207 & 0.58 $\pm$ 0.106 \\
& & \textit{Max - Min} & 0.919 - 0.283 & 0.919 - 0.283 & 0.788 - 0.503 \\
\cline{2-6}
& {\textbf{v} $(\uparrow)$} & \textit{Mean \& Std} & 0.496 $\pm$ 0.194 & 0.461 $\pm$ 0.204 & 0.592 $\pm$ 0.117 \\
& & \textit{Max - Min} & 0.814 - 0.151 & 0.814 - 0.151 & 0.798 - 0.443 \\
\hline
\multirow{6}{*}{\textbf{CS}} & \multirow{2}{*}{\textbf{Norm}}   & \textit{Mean \& Std}      & 0.853 $\pm$ 0.066 & 0.842$\pm$0.064    & 0.883$\pm$0.061 \\
& & \textit{Max - Min} & 0.935 - 0.671 & 0.919 - 0.671 & 0.935 - 0.788 \\
\cline{2-6}
& \multirow{2}{*}{\textbf{u} $(\rightarrow)$} & \textit{Mean \& Std} & 0.648 $\pm$ 0.189 & 0.668 $\pm$ 0.207 & 0.591 $\pm$ 0.101 \\
& & \textit{Max - Min} & 0.95 - 0.309 & 0.95 - 0.309 & 0.79 - 0.513 \\
\cline{2-6}
& \multirow{2}{*}{\textbf{v} $(\uparrow)$} & \textit{Mean \& Std} & 0.508 $\pm$ 0.197 & 0.474 $\pm$ 0.207 & 0.603 $\pm$ 0.122 \\
& & \textit{Max - Min} & 0.829 - 0.156 & 0.829 - 0.156 & 0.798 - 0.44 \\
\hline
\multirow{6}{*}{\textbf{RMSE}} & \multirow{2}{*}{\textbf{Norm}}   & \textit{Mean \& Std}      & 0.877 $\pm$ 0.171 & 0.911$\pm$0.158    & 0.78$\pm$0.17     \\
& & \textit{Max - Min} & 1.307 - 0.614 & 1.307 - 0.706 & 1.02 - 0.614 \\
\cline{2-6}
& \multirow{2}{*}{\textbf{u} $(\rightarrow)$} & \textit{Mean \& Std} & 0.798 $\pm$ 0.251 & 0.765 $\pm$ 0.275 & 0.892 $\pm$ 0.123 \\
& & \textit{Max - Min} & 1.174 - 0.317 & 1.174 - 0.317 & 0.978 - 0.649 \\
\cline{2-6}
& \multirow{2}{*}{\textbf{v} $(\uparrow)$} & \textit{Mean \& Std} & 0.966 $\pm$ 0.209 & 0.999 $\pm$ 0.219 & 0.873 $\pm$ 0.141 \\
& & \textit{Max - Min} & 1.299 - 0.584 & 1.299 - 0.584 & 1.049 - 0.635 \\
\hline
\multirow{6}{*}{\textbf{MAE}} & \multirow{2}{*}{\textbf{Norm}}   & \textit{Mean \& Std}      & 0.65 $\pm$ 0.1 & 0.679$\pm$0.078    & 0.569$\pm$0.11     \\
& & \textit{Max - Min} & 0.817 - 0.448 & 0.817 - 0.551 & 0.712 - 0.448 \\
\cline{2-6}
& \multirow{2}{*}{\textbf{u} $(\rightarrow)$} & \textit{Mean \& Std} & 0.617 $\pm$ 0.213 & 0.611 $\pm$ 0.24 & 0.633 $\pm$ 0.104 \\
& & \textit{Max - Min} & 1.021 - 0.245 & 1.021 - 0.245 & 0.753 - 0.453 \\
\cline{2-6}
& \multirow{2}{*}{\textbf{v} $(\uparrow)$} & \textit{Mean \& Std} & 0.743 $\pm$ 0.187 & 0.786 $\pm$ 0.19 & 0.622 $\pm$ 0.11 \\
& & \textit{Max - Min} & 1.048 - 0.433 & 1.048 - 0.433 & 0.769 - 0.433 \\
\hline
\end{tabular}
\end{table}

\begin{table}[t]
\label{tab:phantom}
\caption{Validation results for phantom environment during the different experiments described in Section \ref{sub:experiments}.}
\small
\hspace{1.5em}
\centering

\begin{tabular}{|l|l|l|cc|cc|}
\hline

\multicolumn{7}{|c|}{\textbf{\large{Phantom}}} \\
\hline
 \multicolumn{3}{|c}{\multirow{2}{*}{\textbf{Experiment}}} & \multicolumn{2}{|c}{\textbf{Static}}  & \multicolumn{2}{|c|}{\textbf{Pumping}} \\
 \cline{4-7}
\multicolumn{3}{|l|}{} & \textbf{Normal} & \textbf{Shear} & \textbf{Normal} & \textbf{Shear} \\
\hline
\multirow{6}{*}{\textbf{CC}} & \multirow{2}{*}{\textbf{Norm}} & \textit{Mean \& Std} & 0.87 $\pm$ 0.02 & 0.75 $\pm$ 0.06 & 0.81 $\pm$ 0.04 & 0.79 $\pm$ 0.01 \\
& & \textit{Max - Min} & 0.90 - 0.83 & 0.81 - 0.67 & 0.87 - 0.78 & 0.79 - 0.78 \\
& \multirow{2}{*}{\textbf{u} $(\rightarrow)$} & \textit{Mean \& Std} & 0.72 $\pm$ 0.23 & 0.53 $\pm$ 0.19 & 0.71 $\pm$ 0.08 & 0.51 $\pm$ 0.10 \\
& & \textit{Max - Min} & 0.92 - 0.28 & 0.75 - 0.28 & 0.82 - 0.62 & 0.61 - 0.41 \\
& \multirow{2}{*}{\textbf{v} $(\uparrow)$} & \textit{Mean \& Std} & 0.56 $\pm$ 0.19 & 0.37 $\pm$ 0.16 & 0.39 $\pm$ 0.24 & 0.42 $\pm$ 0.12 \\
& & \textit{Max - Min} & 0.81 - 0.35 & 0.51 - 0.15 & 0.72 - 0.20 & 0.54 - 0.30 \\
\hline
\multirow{6}{*}{\textbf{CS}} & \multirow{2}{*}{\textbf{Norm}} & \textit{Mean \& Std} & 0.89 $\pm$ 0.02 & 0.78 $\pm$ 0.09 & 0.84 $\pm$ 0.04 & 0.81 $\pm$ 0.02 \\
& & \textit{Max - Min} & 0.92 - 0.85 & 0.88 - 0.67 & 0.91 - 0.81 & 0.82 - 0.79\\
& \multirow{2}{*}{\textbf{u} $(\rightarrow)$} & \textit{Mean \& Std} & 0.75 $\pm$ 0.23 & 0.54 $\pm$ 0.17 & 0.75 $\pm$ 0.09 & 0.51 $\pm$ 0.10 \\
& & \textit{Max - Min} & 0.95 - 0.31 & 0.75 - 0.33 & 0.86 - 0.65 & 0.61 - 0.40 \\
& \multirow{2}{*}{\textbf{v} $(\uparrow)$} & \textit{Mean \& Std} & 0.57 $\pm$ 0.20 & 0.39 $\pm$ 0.17 & 0.41 $\pm$ 0.24 & 0.42 $\pm$ 0.11 \\
& & \textit{Max - Min} & 0.83 - 0.35 & 0.51 - 0.16 & 0.75 - 0.20 & 0.53 - 0.31 \\
\hline
\multirow{6}{*}{\textbf{RMSE}} & \multirow{2}{*}{\textbf{Norm}} & \textit{Mean \& Std} & 0.80 $\pm$ 0.05 & 1.07 $\pm$ 0.20 & 0.90 $\pm$ 0.11 & 1.02 $\pm$ 0.04\\
& & \textit{Max - Min} & 0.86 - 0.71 & 1.31 - 0.82 & 0.99 - 0.75 & 1.05 - 0.98 \\
& \multirow{2}{*}{\textbf{u} $(\rightarrow)$} & \textit{Mean \& Std} & 0.64 $\pm$ 0.32 & 0.94 $\pm$ 0.19 & 0.70 $\pm$ 0.13 & 0.99 $\pm$ 0.10 \\
& & \textit{Max - Min} & 1.17 - 0.32 & 1.16 - 0.70 & 0.83 - 0.53 & 1.09 - 0.88 \\
& \multirow{2}{*}{\textbf{v} $(\uparrow)$} & \textit{Mean \& Std} & 0.90 $\pm$ 0.23 & 1.09 $\pm$ 0.15 & 1.06 $\pm$ 0.24 & 1.07 $\pm$ 0.10 \\
& & \textit{Max - Min} & 1.13 $\pm$ 0.58 & 1.30 - 0.99 & 1.26 - 0.71 & 1.17 - 0.97 \\
\hline
\multirow{6}{*}{\textbf{MAE}} & \multirow{2}{*}{\textbf{Norm}} & \textit{Mean \& Std} & 0.63 $\pm$ 0.05 & 0.71 $\pm$ 0.08 & 0.70 $\pm$ 0.06 & 0.77 $\pm$ 0.05 \\
& & \textit{Max - Min} & 0.68 - 0.55 & 0.79 - 0.60 & 0.74 - 0.61 & 0.82 - 0.72 \\
& \multirow{2}{*}{\textbf{u} $(\rightarrow)$} & \textit{Mean \& Std} & 0.52 $\pm$ 0.26 & 0.78 $\pm$ 0.22 & 0.54 $\pm$ 0.10 & 0.74 $\pm$ 0.11 \\
& & \textit{Max - Min} & 0.96 - 0.25 & 1.02 - 0.48 & 0.64 - 0.41 & 0.85 - 0.63 \\
& \multirow{2}{*}{\textbf{v} $(\uparrow)$} & \textit{Mean \& Std} & 0.69 $\pm$ 0.19 & 0.86 $\pm$ 0.14 & 0.86 $\pm$ 0.21 & 0.85 $\pm$ 0.10 \\
& & \textit{Max - Min} & 0.92 - 0.43 & 1.05 - 0.73 & 1.02 - 0.57 & 0.95 - 0.75 \\
\hline

\end{tabular}
\end{table}

\begin{table}[t]
\label{tab:ex-vivo}
\caption{Validation results for ex-vivo environment during the different experiments described in Section \ref{sub:experiments}.}
\hspace{1.5em}
\small
\centering

\begin{tabular}{|l|l|l|cc|c|}
\hline
\multicolumn{6}{|c|}{\large{\textbf{Ex-vivo}}} \\
\hline
 \multicolumn{3}{|c}{\multirow{2}{*}{\textbf{Experiment}}} & \multicolumn{2}{|c|}{\textbf{Static}}  & \textbf{Pumping} \\
 \cline{4-6}
\multicolumn{3}{|l|}{} & \textbf{Normal} & \textbf{Shear} & \textbf{Normal} \\
\hline
\multirow{6}{*}{\textbf{CC}} & \multirow{2}{*}{\textbf{Norm}} & \textit{Mean \& Std} & 0.93 $\pm$ 0.01 & 0.75 $\pm$ 0.01 & 0.79 $\pm$ 0.01   \\
& & \textit{Max - Min} & 0.94 - 0.93 & 0.76 - 0.74 & 0.80 - 0.78 \\
& \multirow{2}{*}{\textbf{u} $(\rightarrow)$} & \textit{Mean \& Std} & 0.62 $\pm$ 0.12 & 0.50 $\pm$ 0.01 & 0.53 $\pm$ 0.01  \\
& & \textit{Max - Min} & 0.79 - 0.52 & 0.51 - 0.49 & 0.54 - 0.52  \\
& \multirow{2}{*}{\textbf{v} $(\uparrow)$} & \textit{Mean \& Std} & 0.63 $\pm$ 0.12 & 0.62 $\pm$ 0.01 & 0.44 $\pm$ 0.01 \\
& & \textit{Max - Min} & 0.80 - 0.54 & 0.63 - 0.61 & 0.45 - 0.43 \\
\hline
\multirow{6}{*}{\textbf{CS}} & \multirow{2}{*}{\textbf{Norm}} & \textit{Mean \& Std} & 0.93 $\pm$ 0.01 & 0.835 $\pm$ 0.01 & 0.79 $\pm$ 0.01 \\
& & \textit{Max - Min} & 0.94 - 0.93 & 0.85 - 0.83 & 0.80 - 0.78 \\
& \multirow{2}{*}{\textbf{u} $(\rightarrow)$} & \textit{Mean \& Std} & 0.62 $\pm$ 0.12 & 0.56 $\pm$ 0.01 & 0.53 $\pm$ 0.01 \\
& & \textit{Max - Min} & 0.79 - 0.51 & 0.57 - 0.55 & 0.54 - 0.52 \\
& \multirow{2}{*}{\textbf{v} $(\uparrow)$} & \textit{Mean \& Std} & 0.63 $\pm$ 0.12 & 0.67 $\pm$ 0.01 & 0.44 $\pm$ 0.01 \\
& & \textit{Max - Min} & 0.80 - 0.54 & 0.68 - 0.66 & 0.45 - 0.43 \\
\hline
\multirow{6}{*}{\textbf{RMSE}} & \multirow{2}{*}{\textbf{Norm}} & \textit{Mean \& Std} & 0.64 $\pm$ 0.02 & 0.95 $\pm$ 0.01 & 1.02 $\pm$ 0.01 \\
& & \textit{Max - Min} & 0.67 - 0.61 & 0.96 - 0.94 & 1.03 - 1.01 \\
& \multirow{2}{*}{\textbf{u} $(\rightarrow)$} & \textit{Mean \& Std} & 0.85 $\pm$ 0.15 & 0.94 $\pm$ 0.01 & 0.97 $\pm$ 0.01 \\
& & \textit{Max - Min} & 0.98 - 0.65 & 0.95 - 0.93 & 0.98 - 0.96  \\
& \multirow{2}{*}{\textbf{v} $(\uparrow)$} & \textit{Mean \& Std} & 0.84 $\pm$ 0.14 & 0.81 $\pm$ 0.01 & 1.05 $\pm$ 0.01 \\
& & \textit{Max - Min} & 0.94 - 0.64 & 0.82 - 0.80 & 1.06 - 1.04 \\
\hline
\multirow{6}{*}{\textbf{MAE}} & \multirow{2}{*}{\textbf{Norm}} & \textit{Mean \& Std} & 0.48 $\pm$ 0.04 & 0.71 - 0.01 & 0.69 $\pm$ 0.01 \\
& & \textit{Max - Min} & 0.53 - 0.45 & 0.72 - 0.70 & 0.70 - 0.68 \\
& \multirow{2}{*}{\textbf{u} $(\rightarrow)$} & \textit{Mean \& Std} & 0.57 $\pm$ 0.08 & 0.75 $\pm$ 0.01 & 0.71 $\pm$ 0.01 \\
& & \textit{Max - Min} & 0.65 - 0.45 & 0.76 - 0.74 & 0.72 - 0.70 \\
& \multirow{2}{*}{\textbf{v} $(\uparrow)$} & \textit{Mean \& Std} & 0.56 $\pm$ 0.09 & 0.66 $\pm$ 0.01 & 0.77 $\pm$ 0.01 \\
& & \textit{Max - Min} & 0.65 - 0.43 & 0.67 - 0.65 & 0.78 - 0.76 \\
\hline

\end{tabular}
\end{table}
\subsection{Qualitative analysis}

\begin{figure}[t]
    \centering
    \includegraphics[width=\textwidth]{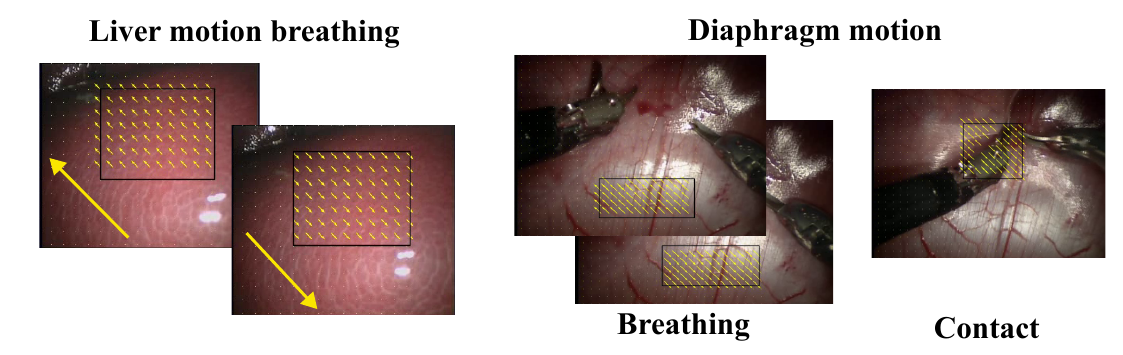}
    \caption{Force texture maps for different real surgical videos. Both contain the analysis of the natural oscillation generated by breathing. On the diaphragm example in the right, we have the analysis of a contact in addition to the underlying breathing motion.}
    \label{fig:qualitative}
\end{figure}

The force textures $\mathcal{F}_{\boldsymbol{\psi}}$ provide the solution of the constraint problem for a certain region of interest inside the image. For plotting purposes, for every time instance $t$ we normalize the force $F_{t}(\boldsymbol{\psi})$ using the spatial mean and standard deviation of the region of interest $\boldsymbol{\psi}$. In addition, to keep the scale across the whole duration $T'$, we normalize the force a second time with the temporal mean and standard deviation. We run our model through all the examples and we validate our method by human inspection and visual consistency with the recorded interaction. Furthermore, we calculate the direction and value of the natural forces from the real surgical videos \cite{mountney_three-dimensional_2010, giannarou_probabilistic_2013, stoyanov_soft-tissue_2005}.  We make a collection with different examples of these textures in Figure \ref{fig:qualitative}. We present the direction and intensity of the forces using yellow arrows and the region of interest is highlighted with a black box.

\section{Discussion}

Our work presents a straightforward approach to the analysis of motion of organs and the surgical force calculation. The use of foundation models demonstrates the possibility to translate such work to multiple scenarios and set ups. Nevertheless, we may observe better results if we overfit the optical flow estimator model to the analysed setup, on the drawback of losing the generalization capability. 

The quantitative results show remarkable matching between the predicted and the force sensor readings. The example presented in Figure \ref{fig:force-comparison} shows that our method can detect and classify the different steps of the interaction. Additionally, it is capable of learning the normalized distribution and range of forces, if the calculated constraint forces are processed using a specific methodology. In terms of the metrics with our whole dataset, we can determine that the model can predict various modes of contact forces such normal pushing force and shear forces. In addition, most of the error in the cross-correlation metric comes from the resting state, as the effect of light change and shadows have a much higher weight when the tissue is kept static under no external influence. However, there are some videos when the force type of contact is more complex and the model clearly fails, such as the example minimum value cross-correlation of the phantom setup presented in the Table \ref{tab:metrics}.

During our analysis, we discover that our method performs best for the ex-vivo environment, even though the difference is not significant. This may be caused by the optical flow estimation model, as the real tissue has a larger variety of features and visual textures to infer from. The absolute value of the optimal lag of the signal, $k$ from Equation \ref{eq:cross-corr}, is kept in a sensible range of $10-30$ which is under a second, that is probably cause by the human error during the execution of the recording scripts.  Nonetheless, based on the statistics and considering that signals with a correlation higher than $0.7$ are largely related, we think that our method provides a sensible estimation of the contact force based on the motion model of tissue for both environments. 

The qualitative analysis also provide promising results. We can determine force not only as a single point contact like most of the previous vision-based methods, but as a texture map that can be displayed on top of the original video. The force is time and direction consistent for the three different scenarios: silicone phantom, ex-vivo tissue and real surgical videos. In addition, we can determine the natural forces acting on the body and remove their effect from the original force sensor and force prediction signal.

Nevertheless, this work presents certain limitations derived from the assumptions we made at the theoretical derivation of our method. Due to weak perspective, we are assuming that the motion towards the camera are small and that can be ignored when projecting into the image space. Therefore, we make the same hypothesis about the forces, large and independent forces towards the camera could cause incorrect predictions of the force. Although this approximation is common in computer vision problems, we could consider adding depth information to account for these extreme cases. In order to match our foundation model based methodology, we could use a deep learning based depth estimator like MiDaS \cite{Ranftl2020, Ranftl2021} and unwarp the object 3D point cloud. 

In addition, during the projection of forces we also loose the physical units of the force. However, as this work is focused onto the generation of haptic feedback, we believe that the real value of the force is not as informative as the force temporal change. As every surgeon experiences force differently, we can always add a user selected scaling factor in order to deploy a sensible feedback for them. In consequence, we assume that this limitation does not directly affect the user case application. Although in this work we lack from the analysis of specific use cases of the real time generation of haptic and visual feedback, we feel that this could be the natural next step of this research.  

Last but not least, we are limited by the user defined binary object mask. In our cases, we manually segmented the organ from a few first frames and use these for all the procedure. In an effort to improve the accuracy of the segmentation, we could use automatized and state-of-the-art deep learning approaches such as Segment Anything \cite{kirillov2023segany}. Moreover, we could use this models to generate more complex scenarios and split the motion basis of each of the elements on the scene. A potential direction to explore could be the use of our methodology to analyse and categorize complex scenarios, that contain multiple

\section{Conclusion}

In this paper, we present a new methodology to predict image-space surgical forces based on a two step learning process: 1) determining the basis of motion from our system using natural motions and 2) estimate the force applied based on the constraint problem. We test our approach in a variety of environments and natural motions under both quantitative and qualitative analysis. During the quantitative analysis our method present promising results in the estimation of contact forces on various contact modes. Event though this research paper presents some limitations, weak perspective or the lack of the absolute physical units; we think that the contributions outweigh them and we demonstrate a robust and reliable pipeline for the given task. To conclude, we believe that this work provides a general theoretical baseline for vision-based force estimation over which new methodologies, deep learning methods and mechanical approaches could potentially be built.

\section*{Funding}

This work was supported by UK Research and Innovation in the Engineering and Physical Sciences Research Council (EPSRC) iCASE [2446549] and the Centre for Artificial Intelligence and Robotics, Hong Kong Institute of Science \& Innovation, Chinese Academy of Sciences [InnoHK].



\bibliographystyle{elsarticle-num} 
\bibliography{bibliography}

\end{document}